\documentclass[sigconf]{acmart}
\AtBeginDocument{%
  }

\copyrightyear{2026}
\acmYear{2026}
\setcopyright{cc}
\setcctype{by}
\acmConference[MM '26]{Proceedings of the 34th ACM International Conference on Multimedia}{November 10--14, 2026}{Rio de Janeiro, Brazil}
\acmBooktitle{Proceedings of the 34th ACM International Conference on Multimedia (MM '26), November 10--14, 2026, Rio de Janeiro, Brazil}
\acmDOI{10.1145/3767308.3837690}
\acmISBN{979-8-4007-2213-4/2026/11}

\usepackage{stfloats}
\usepackage{array}
\usepackage{makecell}
\usepackage{balance}
\usepackage{hyperref}
\begin{document}

\title{KeyID: Decoupled Drafting and Keyframe Editing for Identity-Preserving Video Generation}

\author{Jianjie Luo}
\affiliation{%
  \institution{Guangdong University of Technology}
  \city{Guangzhou}
  \country{China}}
\email{jianjieluo@gdut.edu.cn}

\author{Yiming Zhong}
\affiliation{%
  \institution{Guangdong University of Technology}
  \city{Guangzhou}
  \country{China}}
\email{gdutzym@gmail.com}

\author{Haoming Shen}
\affiliation{%
  \institution{Guangdong University of Technology}
  \city{Guangzhou}
  \country{China}}
\email{3124004329@mail2.gdut.edu.cn}

\author{Yupeng Xiao}
\affiliation{%
  \institution{Guangdong University of Technology}
  \city{Guangzhou}
  \country{China}}
\email{xiaoyupeng1@mails.gdut.edu.cn}

\author{Zhenguo Yang}
\authornote{Corresponding author.}
\affiliation{%
  \institution{Guangdong University of Technology}
  \city{Guangzhou}
  \country{China}}
\email{yzg@gdut.edu.cn}

\renewcommand{\shortauthors}{Jianjie Luo, Yiming Zhong, Haoming Shen, Yupeng Xiao, and Zhenguo Yang}

\begin{abstract}
Identity-preserving video generation (IPVG) requires synthesizing videos that are faithful to both reference subjects and text prompts. Existing methods are often hindered by high tuning costs or limited input-level enhancements, struggling to maintain rigid identity consistency during complex, long-sequence actions. To address these limitations, we propose \textbf{KeyID}, a training-free IPVG framework that decouples the synthesis of video dynamics from the injection of identity. Specifically, KeyID comprises two components: (1) \textit{Reference-Aware Video Generation}, which produces an identity-agnostic video draft aligned with multiple references, and (2) \textit{Identity-Preserved Keyframe Editing}, which integrates the target identity via sparse keyframe correction and subsequent motion interpolation. By shifting from dense frame-level supervision to sparse keyframe-level refinement, KeyID effectively resolves the capacity conflict between prompt adherence and identity fidelity. Crucially, our modular design allows seamless extension to multi-subject references and complex sequential action generation without additional training. KeyID outperforms prior works and is validated by automatic and human evaluations on the official challenge benchmark, ultimately securing the runner-up position in the Track 2 (Sequential Action) of the ACM Multimedia 2026 IPVG Grand Challenge. Source code is available at \url{https://github.com/WISLab-GDUT/KeyID}.
\end{abstract}



%% the work being presented.

\begin{teaserfigure}
  \centering
  \includegraphics[width=0.98\textwidth]{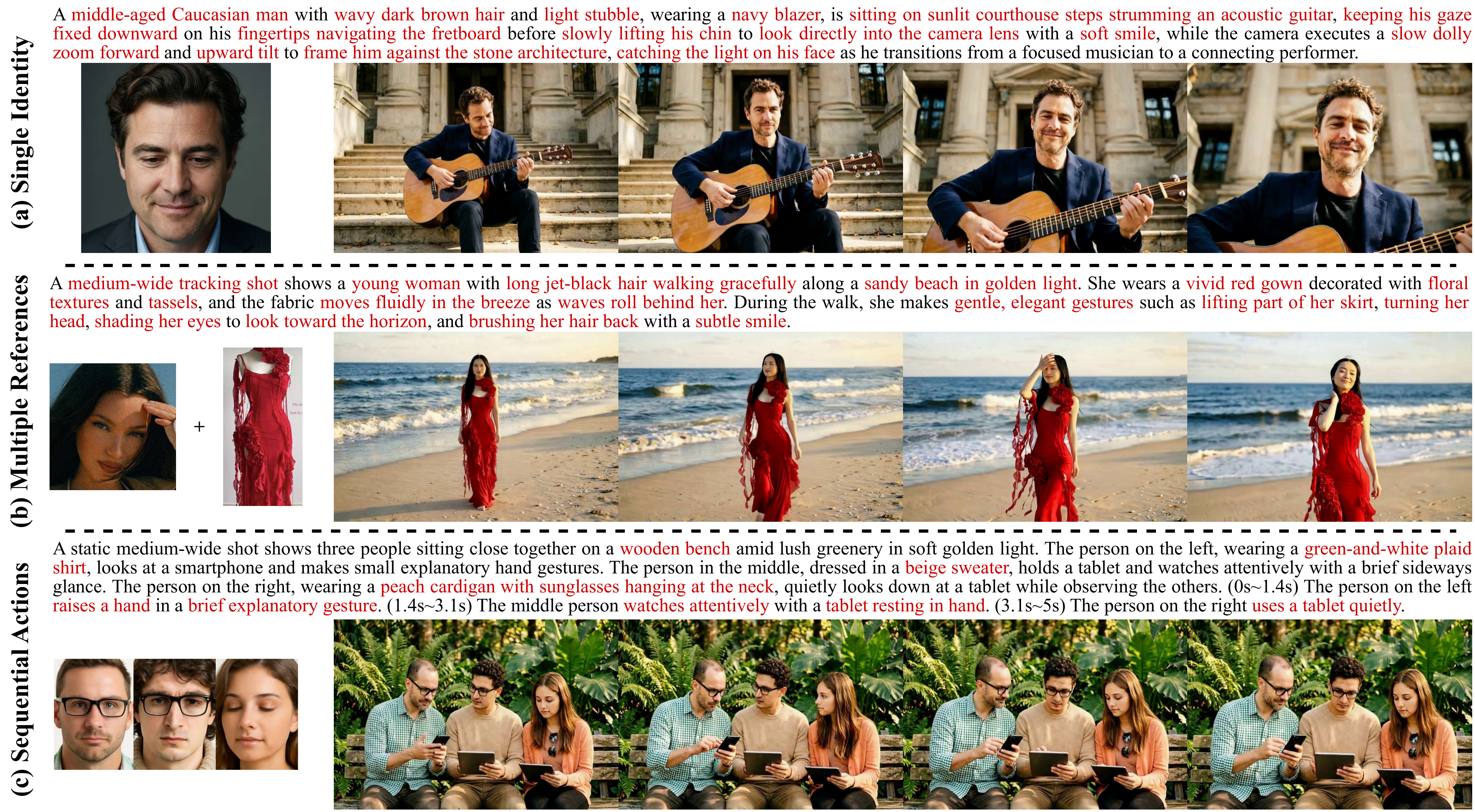}
  \vspace{-0.15in}
  \caption{KeyID generates higher-quality identity-preserving videos from (a) Single Identity, (b) Multiple References, and (c) Sequential Actions. \textcolor{red}{Red} text highlights key attributes and actions in long instructions.
  }
  \label{fig:teaser}
\end{teaserfigure}

\maketitle

\section{Introduction}

Identity-preserving video generation (IPVG)~\cite{yuan2025identity, zhang2025fantasyid} aims to synthesize temporally coherent videos from text prompts while maintaining reference identities. While early IPVG methods focus on single-identity and single-prompt settings (Fig.~\ref{fig:teaser}(a)), practical applications~\cite{wei2024dreamvideo, chen2504skyreels, guo2026dreamid2} now require more complex scenarios. For instance, modern generation applications often involve multi-reference generation (Fig.~\ref{fig:teaser}(b)), which integrates diverse visual concepts, such as additional reference objects, without letting one reference dominate the others~\cite{zhong2025concat, chen2025humo, wang2026refalign, lireactid}. They also require sequential action control (Fig.~\ref{fig:teaser}(c)), where models must execute specific motion transitions based on timestamped prompts (e.g., ``0--1.4s: raises a hand'')~\cite{lireactid, chen2026prompt}. Thus, the current objective of IPVG is to generate long-sequence actions while preserving strict identity consistency.

However, existing methods often struggle to satisfy these advanced requirements due to a fundamental semantic gap between the facial reference and the text prompt. This inherent gap forces the generator to balance prompt adherence and identity fidelity within a single video generation process. One paradigm attempts to bridge this gap by fine-tuning large pre-trained video diffusion models on ID-matched data~\cite{li2025personalvideo, zhang2025magicmirror}. Although this approach achieves impressive results, severe data scarcity and prohibitive tuning costs prevent the base video generation models from scaling effectively and sufficiently adapting to the IPVG task. Therefore, a subsequent paradigm explores training-free approaches~\cite{gao2025identity, wang2025identity} through input-level enhancements. Nevertheless, because both paradigms still rely on an end-to-end generation process, they fail to resolve the underlying trade-off. Consequently, these methods remain highly susceptible to temporal identity drift, struggling to maintain the fine-grained facial characteristics of the target identity across long-sequence actions. Ultimately, the demand for multi-reference generation and precise sequential actions within such a coupled generation process overwhelms the model, rendering robust identity preservation exceedingly difficult.

To address the limitations of coupled generation, we propose \textbf{KeyID}, a training-free IPVG framework that decouples the synthesis of video dynamics from the injection of identity. Specifically, KeyID reformulates identity preservation as sparse keyframe correction followed by motion interpolation, which comprises two primary components: \textit{\textbf{R}eference-\textbf{A}ware \textbf{V}ideo \textbf{G}eneration} (\textbf{RAVG}) and \textit{\textbf{I}dentity-\textbf{P}reserved \textbf{K}eyframe \textbf{E}diting} (\textbf{IPKE}). Firstly, RAVG produces an identity-agnostic, spatiotemporally consistent video draft that aligns with multiple references, including the text prompt and general objects. This is achieved through prompt enhancement and first-frame editing, followed by multi-event draft expansion. Secondly, IPKE integrates the target identity into the draft via identity-aware video editing. It extracts a compact set of representative keyframes from the video draft and applies explicit identity refinement to each keyframe through face swapping. Finally, the complete output video is reconstructed by interpolating the intermediate motions between these corrected keyframes.

In sum, we have made the following contributions: 
(\textbf{I}) KeyID establishes a novel paradigm for identity-preserving video generation that decouples the synthesis of video dynamics from identity integration. 
(\textbf{II}) This decoupled framework offers an effective solution for complex generation scenarios, particularly those requiring multi-reference conditions and sequential actions.
(\textbf{III}) The proposed method has been validated in the ACM Multimedia 2026 IPVG Grand Challenge~\cite{panidentity}, where it surpasses the first-place result of the previous year in the Track 1 (Facial) setting and secures the runner-up position in the Track 2 (Sequential Action) setting.

\section{Related Work}

\textbf{Text-to-video Diffusion Models.}
Diffusion models~\cite{ho2020denoising, rombach2022high, esser2024scaling, gao2025conmo} have become the dominant paradigm for visual generation. Early text-to-video (T2V) models~\cite{chefer2024still,wei2025echovideo,zhang2025motion}, such as VideoCrafter~\cite{chen2023videocrafter1}, ModelScope~\cite{wang2023modelscope}, and Stable Video Diffusion~\cite{blattmann2023stable}, established the capability of generating temporally coherent videos from text prompts. More recently, Diffusion Transformer~\cite{peebles2023scalable} backbones have become the standard architecture. Models such as CogVideoX~\cite{yang2025cogvideox}, HunyuanVideo~\cite{kong2024hunyuanvideo}, and Wan~\cite{wan2025wan} demonstrate strong scalability and high generation quality. VACE~\cite{jiang2025vace} further supports multi-input conditioning for unified video generation and editing. While these foundation models excel at general conditional video generation, they fall short on the specific demands of identity preservation.

\textbf{Training-based IPVG.}
A common paradigm augments foundation models with extra encoders or adapters fine-tuned on ID-matched data for identity preservation~\cite{he2024id,huang2025conceptmaster,ye2023ip,zhang2025magicmirror}. For instance, ConsisID~\cite{yuan2025identity} utilizes frequency decomposition to separate identity features from motion dynamics. Concat-ID~\cite{zhong2025concat} concatenates image features with video latents through 3D self-attention. Stand-In~\cite{xue2026stand} integrates identity features via LoRA~\cite{hu2022lora} and restricted self-attention. RefAlign~\cite{wang2026refalign} extends this approach to multi-object preservation through fine-tuning of the reference branch. ReactID~\cite{lireactid} extends IPVG to sequential actions through action-conditioned sub-tasks. However, fine-tuning on scarce ID-matched data weakens prompt adherence, especially under sequential action prompts demanding precise motion control.

\textbf{Training-free IPVG.}
To bypass data scarcity and reduce tuning costs, training-free methods enhance conditional input to maximize native model capability without modifying foundation model parameters. For example, TPIGE~\cite{gao2025identity} applies face-aware prompt enhancement and identity-aware spatiotemporal guidance. Wang et al.~\cite{wang2025identity} decompose the task into cascaded text-to-image and image-to-video generation. While effective at reducing tuning costs, these pre-processing methods enhance inputs without subsequent correction, leaving them vulnerable to identity drift. KeyID addresses this through a post-processing pipeline: it first generates a video draft from the reference and prompt, then refines identity via sparse keyframe editing followed by motion interpolation, naturally extending to multi-reference and sequential action settings.

\section{Method}

\begin{figure*}[tb]
    \centering
    \includegraphics[width=\textwidth]{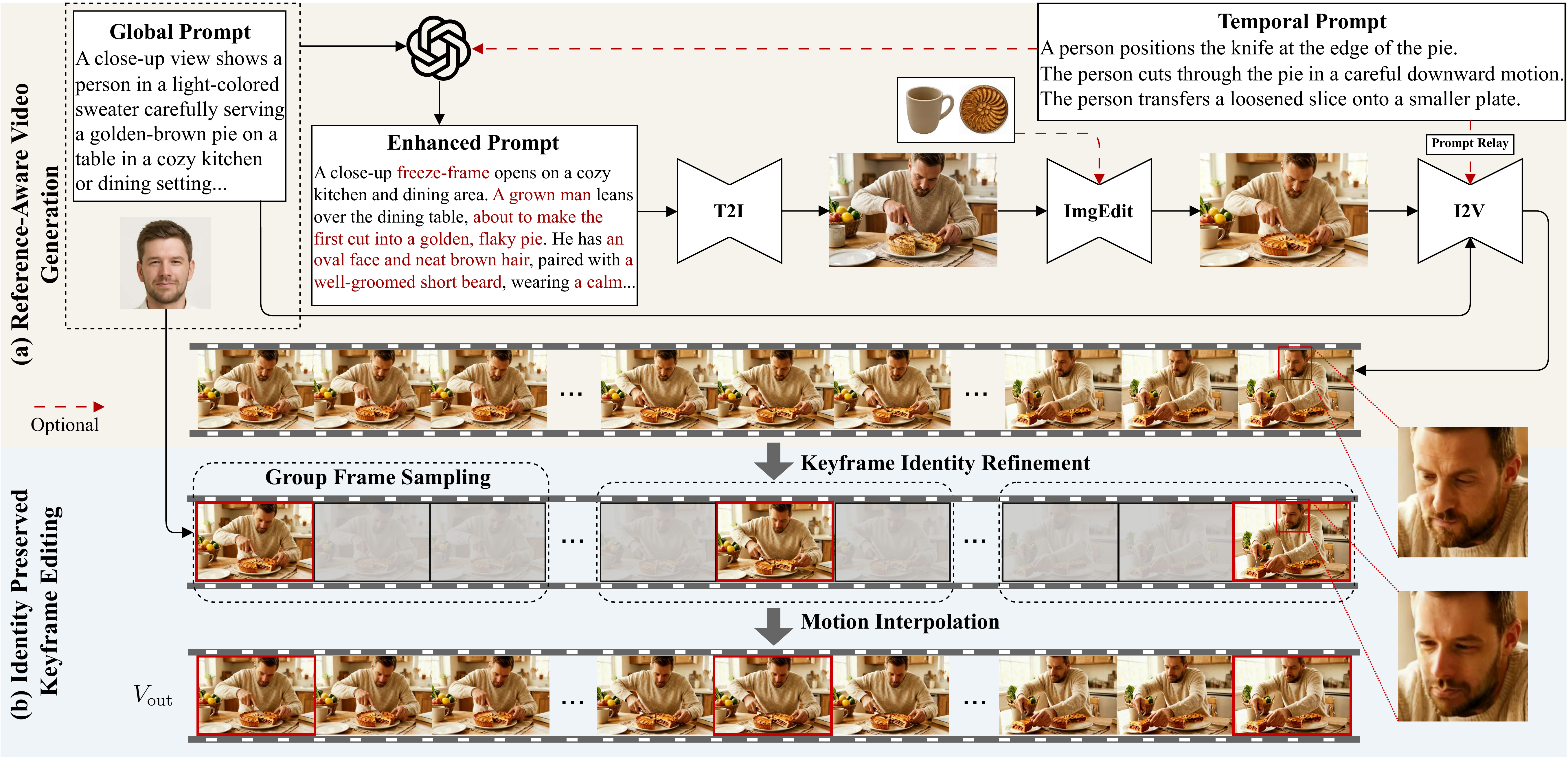}
    \caption{Overview of KeyID. (a) Reference-Aware Video Generation. A global prompt, optional temporal prompts, and a face reference are fused by GPT-5 into an enhanced prompt, from which a first frame is generated and optionally edited with non-human objects. The edited frame and the global prompt, along with optional temporal prompts, drive I2V to produce a video draft. (b) Identity Preserved Keyframe Editing. Keyframe windows are uniformly sampled in groups along the video draft, optimal keyframes are selected in keyframe identity refinement, and motion interpolation produces the final temporally coherent video with preserved identity and prompt adherence.}
    \label{fig:framework}
\end{figure*}

\subsection{Overview}

Given a reference identity image $I_{\text{ref}}$, a set of $N_o$ reference images for non-human objects $\mathcal{I}_{\text{obj}} = \{I_{\text{obj}}^i\}^{N_o}_{i=1} $, and a global prompt $P_0$, the objective of IPVG is to generate a video $V_{\text{out}}$ consisting of $M$ frames over $K$ seconds, in which the reference identity is faithfully maintained and the reference objects appear naturally within the scene. For the sequential action IPVG, the input additionally includes a set of $N_t$ timestamped segment prompts $\mathcal{P}_{\text{tem}} = \{(P_i, t_{i-1}, t_i)\}_{i=1}^{N_t}$, each describing the action for the corresponding time interval.

The core design principle of KeyID is to decouple the synthesis of video dynamics from the identity integration. Specifically, KeyID comprises two primary components: (1) \textit{\textbf{R}eference-\textbf{A}ware \textbf{V}ideo \textbf{G}eneration} (\textbf{RAVG}) and (2) \textit{\textbf{I}dentity-\textbf{P}reserved \textbf{K}eyframe \textbf{E}diting} (\textbf{IPKE}). As illustrated in Figure~\ref{fig:framework}, RAVG first generates an identity-agnostic video draft $V_{\text{draft}}$ of $N$ frames that aligns with the multi-reference conditions. This strategy enables the video generation foundation model to allocate full capacity toward prompt adherence and temporal coherence. Subsequently, IPKE integrates the target identity into the draft by performing sparse keyframe correction and interpolating the intermediate motions to produce the final output video $V_{\text{out}}$. To support the sequential action IPVG task, RAVG can optionally be augmented with Prompt Relay~\cite{chen2026prompt}, which serves as an inference-time temporal routing module that assigns each timeline segment to the appropriate action description.

\subsection{Reference-Aware Video Generation}

RAVG produces a timeline-aligned video draft without imposing explicit identity constraints during generation. By deferring identity preservation to IPKE, the video generation foundation model operates at full capacity for prompt following and temporal coherence. Technically, RAVG consists of three steps: Prompt Enhancement, First-frame Generation, and Video Draft Generation.

\textbf{Prompt Enhancement.}
Standard video prompts $P_0$ often lack the fine-grained visual details (e.g., age, hairstyles, facial hair) required for high-quality image generation.
To bridge this gap, we use GPT-5~\cite{singh2025openai} to extract a concise subject description $P_s$ from $I_{\text{ref}}$. We then fuse $P_s$ with $P_0$ and $\mathcal{P}_{\text{tem}}$ to produce an enhanced prompt $\bar{P}$. This step ensures that the subsequent T2I process has sufficient descriptive prior to generate a subject that matches the reference's non-facial attributes (e.g., hair, clothing) which are not explicitly handled by face-swapping. 

\textbf{First-frame Generation.}
The first frame defines the spatial layout and initial identity. We generate an initial first frame from the enriched prompt $\bar{P}$ via a T2I model:
\begin{equation}
I_1 = \text{T2I}(\bar{P}),
\end{equation}
However, text prompts are often insufficient to describe complex non-human object identities (e.g., a specific pink basketball with unique textures in $\mathcal{I}_{\text{obj}}$). Thus, we apply an ImgEdit model to composite these objects:
\begin{equation}
\bar{I}_1 = \text{ImgEdit}(I_1, \mathcal{I}_{\text{obj}}).
\end{equation}
By injecting object identities at the image level, we ensure the I2V model starts with a complete visual reference for both the person and the objects. Notably, this detailed anchor frame reduces the generative burden on the video model, thereby allowing it to focus entirely on motion dynamics.

\textbf{Video Draft Generation.} 
Using $\bar{I}_1$ as the starting condition, we generate an identity-agnostic video draft using an I2V model:
\begin{equation}
V_{\text{draft}} = \{\bar{I}_1, I_2, \dots, I_{N}\} = \text{I2V}(\bar{I}_1, P_0, \mathcal{P}_{\text{tem}}).
\end{equation}
While non-human object identities are relatively stable under standard I2V diffusion, human facial identity remains prone to drift. By deliberately deferring this facial correction to the subsequent stage, the I2V model is freed to focus entirely on complex motion synthesis. Consequently, to seamlessly handle sequential actions without semantic entanglement, we augment the I2V process with Prompt Relay~\cite{chen2026prompt}. This module introduces a temporal routing penalty in the cross-attention layers, ensuring that frames within a specific interval $(t_{i-1}, t_i)$ attend primarily to their corresponding action description $P_i$. The resulting $V_{\text{draft}}$ provides the motion backbone for the final output.

\subsection{Identity Preserved Keyframe Editing}

While RAVG ensures motion coherence, the facial identity in $V_{\text{draft}}$ may still deviate from $I_{\text{ref}}$. The motivation for IPKE is to apply explicit identity correction without introducing temporal flickering. Since per-frame editing frequently causes spatiotemporal jitter, we reformulate identity preservation as sparse keyframe refinement followed by motion-aware interpolation.

\textbf{Group Frame Sampling.}
Not all frames in $V_{\text{draft}}$ are equally suitable for identity refinement. Motion blur and extreme poses can cause face-swapping to fail. To balance temporal coverage with content quality, we distribute $K$ nominal keyframe positions $k_i$ across the video. For each $k_i$, we define a local search window:
\begin{equation}
W_i = [k_i - w, k_i + w] \cap [1, N].
\end{equation}
This redundancy allows us to bypass degraded frames and select the optimal candidate for refinement.

\textbf{Keyframe Identity Refinement.}
For every candidate $I_j$ in $W_i$, we perform explicit Identity Refinement using a FaceSwap model\footnote{\url{https://github.com/deepinsight/insightface}}:
\begin{equation}
I_j^{\text{ref}} = \text{FaceSwap}(I_j, I_{\text{ref}}).
\end{equation}
We then use ArcFace~\cite{deng2019arcface} to compute a similarity score $s_j$ against the reference. The final keyframe for the $i$-th window is selected as:
\begin{equation}
\bar{k}_i = \arg\max_{j \in W_i} s_j.
\end{equation}
This decoupling is highly advantageous. It allows the use of off-the-shelf, high-resolution faceswap models whose capacity is not shared with the generation model, ensuring near-perfect identity fidelity on the frames that anchor subsequent interpolation.

\textbf{Motion Interpolation.}
To reconstruct the final video $V_{\text{out}}$, we populate the gaps between refined keyframes. Simply concatenating edited frames would lead to discontinuities. The keyframes reside in the $N$-frame draft space, while the output targets $M$ frames over the same duration. We map each $\bar{k}_i$ via linear temporal rescaling:
\begin{equation}
t_i = \left\lfloor \frac{M}{N} \cdot \bar{k}_i \right\rceil.
\end{equation}
Given the refined keyframes $I_{\bar{k}_i}^{\text{ref}}$ at their mapped positions $t_i$, we use LTX-2~\cite{hacohen2026ltx} to synthesize the intermediate frames. All keyframes are fed simultaneously, and the model performs motion-compensated interpolation across the full sequence in one pass:
\begin{equation}
V_{\text{out}} = \text{LTX-2}\bigl(I_{t_1}^{\text{ref}}, \dots, I_{t_K}^{\text{ref}}\bigr),
\end{equation}
where $K$ denotes the number of keyframes. By conditioning interpolation on these identity-corrected anchors, LTX-2 propagates high-fidelity identity across the full sequence, producing a temporally smooth video of $M$ frames.

\section{Experiments}

\subsection{Dataset and Evaluation Metrics}

\textbf{Datasets.}
To evaluate KeyID, we conduct experiments on the Facial and Sequential Action Tracks of the ACM Multimedia 2026 Identity-Preserving Video Generation Challenge\footnote{\url{https://hidream-ai.github.io/ipvg-challenge-2026.github.io/}}. As Facial IPVG (Track~1) is foundational to the complex Sequential Action IPVG (Track~2), we use the official VIP-200K test dataset~\cite{zhang2025identity} for algorithm tuning and capability validation. Specifically, this test dataset contains 200 unseen person IDs, with each ID featuring portrait images and five textual prompts for video generation, totaling 1,000 test pairs. For the Sequential Action IPVG, we evaluate KeyID on the ACM MM 2026 IPVG Track~2 dataset. This test set contains 200 samples, each comprising reference identity images, non-human objects, a global prompt, and timestamped action captions.

\textbf{Evaluation Metrics.} 
Following the challenge protocol, the official Track~2 ranking uses a combined score of objective metrics and human evaluation. To provide detailed diagnostics beyond the aggregate score, we also report standard objective metrics following established Facial IPVG practices~\cite{zhang2025identity}. Specifically, Face-Cur and Face-Arc measure identity consistency via CurricularFace~\cite{huang2020curricularface} and ArcFace~\cite{deng2019arcface}. Furthermore, CLIPScore~\cite{hessel2021clipscore} quantifies the semantic alignment between generated videos and text prompts.

\subsection{Implementation Details}
The subject recognition and prompt enhancement modules of KeyID are implemented using GPT-5.4~\cite{singh2025openai}. The initial frame is generated by ERNIE-Image-Turbo\footnote{\url{https://github.com/baidu/ernie-image}}, and Qwen-Image-Edit-2511~\cite{wu2025qwen} is subsequently utilized to inject non-human reference objects into this frame. For the generation of the video draft, Wan2.2~\cite{wan2025wan} is adopted as the image-to-video diffusion backbone, which is augmented with Prompt Relay~\cite{chen2026prompt} to achieve timeline-aware temporal control. During the stage of explicit identity refinement, the similarity of the identity is evaluated by ArcFace~\cite{deng2019arcface}. Additionally, LTX-2.3~\cite{hacohen2026ltx} serves as the model for motion interpolation. Regarding the temporal configuration, the video draft is generated to consist of $N=81$ frames with a duration of 5 seconds. The process of keyframe selection extracts $K=5$ keyframes using a search window radius of $w = 8$. Ultimately, the final output video comprises $M=121$ frames within the same duration.

\begin{table}[tb]
\centering
\caption{Comparison results of KeyID with other state-of-the-art Facial IPVG approaches on the VIP-200K test dataset.}
\label{tab:fine_grained}
\begin{tabular}{lccc}
\toprule
Method                    & Face-Cur$\uparrow$ & Face-Arc$\uparrow$  & CLIPScore$\uparrow$ \\
\midrule
Concat-ID~\cite{zhong2025concat}       & 0.242                          & 0.228                           & \underline{29.0}                \\
Xu et al.~\cite{xu2025improving}       & 0.285                          & 0.269                           & 28.6                            \\
Wang et al.~\cite{wang2025identity}    & 0.467                          & 0.441                           & 28.0                            \\
TPIGE~\cite{gao2025identity}           & \underline{0.492}              & \underline{0.473}               & 27.8                            \\
\midrule
KeyID (Ours)                           & \textbf{0.633}                 & \textbf{0.630}                  & \textbf{30.9}                   \\
\bottomrule
\end{tabular}
\end{table}

\subsection{Performance Comparison}

\begin{table}[tb]
\centering
\caption{Comparison results of KeyID with other state-of-the-art Sequential Action IPVG approaches on the leaderboard of IPVG 2026 Challenge (Track~2).}
\label{tab:leaderboard}
\begin{tabular}{lcc}
\toprule
Team & Final Score$\downarrow$ & Rank \\
\midrule
USTC-CMI                      & 1.25          & 1 \\
\textbf{WislabGDUT (Ours)}    & \textbf{1.81} & \textbf{2} \\
USTC-AC                       & 2.94          & 3 \\
\bottomrule
\end{tabular}
\vspace{-0.1in}
\end{table}

\textbf{Facial IPVG Evaluation.}
Table~\ref{tab:fine_grained} summarizes the results on the VIP-200K test dataset, comparing KeyID with Concat-ID~\cite{zhong2025concat} and the top three methods from IPVG 2025 Challenge~\cite{zhang2025identity}: TPIGE~\cite{gao2025identity}, Wang et al.~\cite{wang2025identity}, and Xu et al.~\cite{xu2025improving}. Since IPVG 2026 submissions are not publicly available, this evaluation provides a controlled comparison with prior representative IPVG methods. KeyID achieves the best identity preservation scores, with Face-Cur of 0.633 and Face-Arc of 0.630, improving over the strongest baseline TPIGE by +28.7\% and +33.2\%, respectively. The consistent gains on Face-based metrics indicate stronger identity fidelity across different face similarity measures. KeyID also obtains the highest CLIPScore of 30.9, suggesting that sparse keyframe-level identity refinement improves identity fidelity while maintaining text-video alignment.

\textbf{Sequential Action IPVG Evaluation.}
Table~\ref{tab:leaderboard} reports the official leaderboard results on Track~2 of the IPVG 2026 Challenge~\cite{panidentity}, which focuses on sequential action identity-preserving video generation. Notably, KeyID achieves a highly competitive final score of 1.81 and secures the runner-up position. This outcome clearly demonstrates the effectiveness of decoupling video dynamics from identity injection to prevent spatial-temporal interference in a significantly more complex setting, where the model must preserve identity fidelity while accurately following timestamped action prompts and incorporating non-human reference objects.

\begin{figure*}[tb]
    \centering
    \includegraphics[width=0.95\textwidth]{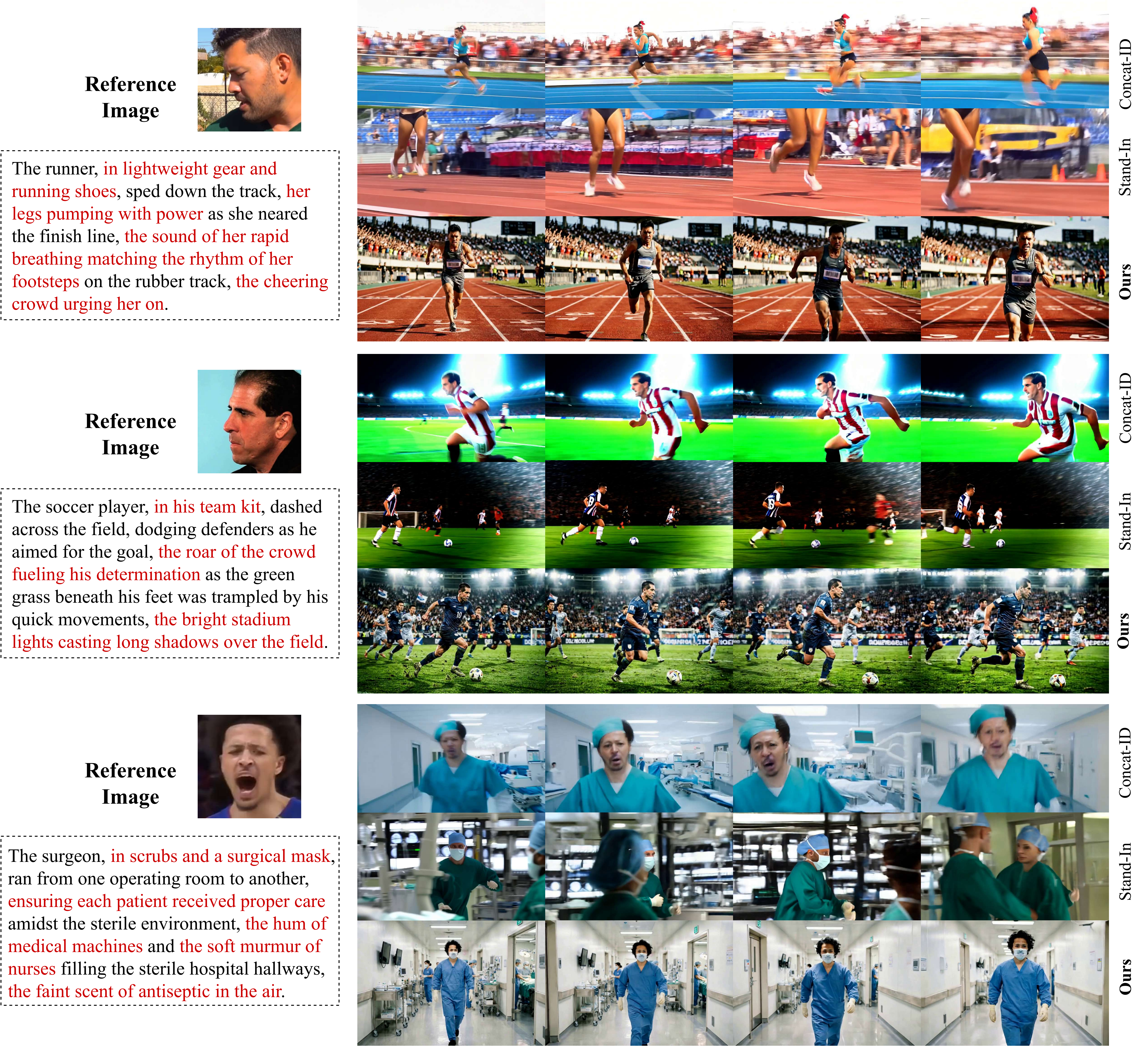}
    \vspace{-0.1in}
    \caption{Qualitative comparison. Four evenly sampled frames are shown for each video. KeyID better preserves identity consistency and prompt adherence across complex actions.}
    \Description{Four frames each from videos generated by KeyID and baselines, showing identity consistency across action segments.}
    \label{fig:qualitative}
\end{figure*}

\textbf{Qualitative Analysis.}
Figure~\ref{fig:qualitative} compares KeyID with two representative identity-preserving video generation baselines, Concat-ID~\cite{zhong2025concat} and Stand-In~\cite{xue2026stand}. These methods are selected because they represent different identity-conditioning strategies from us and provide strong baselines for evaluating identity consistency and prompt adherence. We show four evenly sampled frames for each generated video. Stand-In generates visually appealing frames, but its identity visibility can become unstable under large motion, e.g., in the first case, the generated subject is shown mostly with the lower body and the face region is absent. Concat-ID preserves partial identity resemblance, but it tends to weaken fine-grained prompt adherence in complex scenes, e.g., in the last case, it fails to include the surgical mask specified in the prompt. In contrast, KeyID maintains the reference identity across frames while better following the prescribed actions and visual attributes. These results suggest that sparse keyframe-level identity refinement improves identity fidelity without visibly weakening prompt adherence.

\subsection{Experimental Analysis}

\begin{table}[tb]
\centering
\caption{Ablation study on each component in KeyID on the subset of the VIP-200K test dataset.}
\label{tab:component_breakdown}
\begin{tabular}{>{\centering\arraybackslash}c cccccc}
\toprule
Base & IPKE & RAVG  & Face-Cur$\uparrow$ & Face-Arc$\uparrow$ & CLIPScore$\uparrow$ \\
\midrule
\checkmark    &             &               & 0.279                          & 0.260                          & \underline{28.6}                         \\
\checkmark    &\checkmark   &               & \underline{0.596}              & \underline{0.594}              & 28.3                                     \\
\checkmark    &\checkmark   & \checkmark    & \textbf{0.636}                 & \textbf{0.643}                 & \textbf{30.9}   \\
\bottomrule
\end{tabular}
\end{table}

\begin{table}[tb]
\centering
\caption{Performance comparison of KeyID and different variants on the subset of the VIP-200K test dataset.}
\label{tab:ablation}
\setlength{\tabcolsep}{4pt}
\begin{tabular}{lccc}
\toprule
Configuration & Face-Cur$\uparrow$ & Face-Arc$\uparrow$ & CLIPScore$\uparrow$ \\
\midrule
T2V + IPKE                & 0.410                          & 0.395                          & 29.2                            \\
RAVG + DreamID-V~\cite{guo2026dreamid} 
                          & \underline{0.503}              & \underline{0.484}              & \underline{30.8}                \\
RAVG + IPKE (KeyID)       & \textbf{0.636}                 & \textbf{0.643}                 & \textbf{30.9}                   \\
\bottomrule
\end{tabular}
\vspace{-0.15in}
\end{table}

To accelerate the experimental validation process, we conduct our ablation studies on a subset of 200 samples from the VIP-200K test dataset. Specifically, this subset is systematically formed by sampling exactly one prompt per identity.

\textbf{Ablation Study.} 
Table~\ref{tab:component_breakdown} reports per-component contributions, using Xu et al.~\cite{xu2025improving} as the baseline generation method. Row~1 shows the baseline alone, which achieves low identity scores (Face-Cur 0.279, Face-Arc 0.260). Applying IPKE (Identity Preserved Keyframe Editing) as post-processing on the baseline's output (row~2) yields a sharp identity gain: Face-Cur from 0.279 to 0.596 (+0.317) and Face-Arc from 0.260 to 0.594 (+0.334), demonstrating that IPKE alone substantially improves identity fidelity even without modifying the generation pipeline. Adding RAVG (Reference-Aware Video Generation) further lifts CLIPScore to 30.9, Face-Cur to 0.636, and Face-Arc to 0.643, confirming that RAVG strengthens draft quality and provides a better foundation for IPKE.

\textbf{Impact of Generation and Identity Strategies.}
Table~\ref{tab:ablation} isolates the keyframe-based identity strategy. Replacing RAVG with vanilla T2V (T2V + IPKE) sharply decreases identity scores (Face-Cur $-0.226$, Face-Arc $-0.248$), indicating that RAVG provides a stronger foundation draft for downstream identity refinement. Substituting IPKE with dense video-level identity transfer (RAVG + DreamID-V~\cite{guo2026dreamid}) also leads to substantially lower identity scores than KeyID, with Face-Cur decreasing from 0.636 to 0.503 and Face-Arc from 0.643 to 0.484. This comparison shows that sparse keyframe correction is more effective for preserving identity than dense video-level identity transfer. The full model (RAVG + IPKE) achieves the best identity metrics and CLIPScore, suggesting that KeyID improves identity fidelity while retaining prompt adherence.

\section{Conclusion}

In this paper, we propose KeyID, a training-free identity-preserving video generation framework designed to balance prompt adherence and identity fidelity in multi-reference generation and complex action sequences. By decoupling video dynamics from identity injection, KeyID first produces an identity-agnostic video draft and then integrates identity through sparse keyframe correction and motion interpolation. This approach addresses the inherent trade-off between following textual prompts and maintaining subject identity. KeyID achieved second place in Track 2 of the ACM Multimedia 2026 Identity-Preserving Video Generation Grand Challenge, and our evaluations suggest that keyframe-level refinement is essential for high-quality video synthesis.

\bibliographystyle{ACM-Reference-Format}
\balance
\bibliography{references}

% \appendix

\end{document}